\documentclass[runningheads]{llncs}
\usepackage{graphicx}
\usepackage{mdframed}
\usepackage{booktabs}

\usepackage[hyphens]{url}
\usepackage{hyperref}
\usepackage[hyphenbreaks]{breakurl}

\begin{document}
\title{Financial Report Chunking for Effective Retrieval Augmented Generation}
%
%
\author{Antonio Jimeno Yepes \and Yao You \and Jan Milczek \and Sebastian Laverde \and Leah Li}

\authorrunning{Jimeno Yepes et al.}
%
\institute{Unstructured Technologies\\ Sacramento, CA, USA \\
\email{leah@unstructured.io}\\
\url{https://unstructured.io}}

\maketitle              
\begin{abstract}
Chunking information is a key step in Retrieval Augmented Generation (RAG). Current research primarily centers on paragraph-level chunking. This approach treats all texts as equal and neglects the information contained in the structure of documents. We propose an expanded approach to chunk documents by moving beyond mere paragraph-level chunking to chunk primary by structural element components of documents. Dissecting documents into these constituent elements creates a new way to chunk documents that yields the best chunk size without tuning. We introduce a novel framework that evaluates how chunking based on element types annotated by document understanding models contributes to the overall context and accuracy of the information retrieved. We also demonstrate how this approach impacts RAG assisted Question \& Answer task performance. Our research includes a comprehensive analysis of various element types, their role in effective information retrieval, and the impact they have on the quality of RAG outputs. Findings support that element type based chunking largely improve RAG results on financial reporting. Through this research, we are also able to answer how to uncover highly accurate RAG.



\end{abstract}
\section{Introduction}



Existing approaches for document understanding use a combination of methods from the computer vision and natural language processing domains to identify the different components in a document. In the rapidly evolving landscape of artificial intelligence, the capability to effectively process unstructured data is becoming increasingly critical. Large Language Models (LLMs) like GPT-4 have revolutionized natural language understanding and generation, as evidenced by their prompt-based functionalities~\cite{openai2023gpt4}, enabling a wide range of applications~\cite{chen2023chatgpt}. However, the efficacy of these models is often constrained by their reliance on the size and quality of the data they process. A notable limitation is the restricted contextual window of LLMs, which hampers their ability to fully comprehend the contents of extensive documents
~\cite{liu2023lost,longchat2023,kaddour2023challenges}. 
By dissecting large volumes of text into smaller, more focused segments, LLMs can process each part with greater precision, ensuring a thorough understanding of each section. This segmented approach allows for meticulous analysis of unstructured data, enabling LLMs to construct a more comprehensive and coherent understanding of the entire document~\cite{xu2023retrieval}. There remains a challenge in 
ensuring factual accuracy and relevance in the generated responses, especially when dealing with complex or extensive information. 

Recently, Retrieval Augmented Generation (RAG)~\cite{lewis2020retrieval,gao2023retrieval} has been developed to address the~\emph{hallucination} problem with LLMs~\cite{ji2023hallucination,ye2023cognitive} when recovering factual information directly from an LLM. In RAG, instead of answering a user query directly using an LLM, the user query is used to retrieve documents or segments from a corpus and the top retrieved documents or segments are used to generate the answer in conjunction with an LLM.
In this way, RAG constraints the answer to the set of retrieved documents.
RAGs have been used as well to answer questions from single documents~\cite{islam2023financebench}.
The documents are split into smaller parts or chunks, indexed by a retrieval system and recovered and processed depending on the user information need.
In a sense, this process allows answering questions about information in a single document, thus contributing to the set of techniques available for document understanding.

Since documents need to be~\emph{chunked} for RAG processing, this raises the question about what is the best practice to chunk documents for effective RAG document understanding.
There are several dimensions to consider when deciding how to chunk a document, which includes the size of the chunks.

The retrieval system in RAG can use traditional retrieval systems using bag-of-words methods or a vector database. If a vector database is used, then an embedding needs to be obtained from each chunk, thus the number of tokens in the chunk is relevant since the neural networks processing the chunks might have constraints on the number of tokens. As well, different chunk sizes might have undesirable retrieval results.
Since the most relevant retrieved chunks need to be processed by an LLM, the number of tokens in retrieved chunks might have an effect in the generation of the answer~\cite{liu2023lost}.
As we see, chunking is required for RAG systems and there are several advantages and disadvantages when considering how to chunk a document.


In this work, we study specifically the chunking of U.S. Securities and Exchange Commission (SEC)\footnote{\url{https://www.sec.gov}} Financial Reports\footnote{\url{https://www.sec.gov/files/cf-frm.pdf}}, including 10-Ks, 10-Qs, and 8-Ks. This study plays a critical role in offering insights into the financial health and operational dynamics of public companies. These documents present unique challenges in terms of document processing and information extraction as they consist of varying sizes and layouts, and contain a variety of tabular information.
Previous work has evaluated the processing of these reports with simple chunking strategies (e.g., tokens), but we believe that a more effective use of these reports might be achieved by a better pre-processing of the documents and chunking configuration\footnote{\url{https://www.cnbc.com/2023/12/19/gpt-and-other-ai-models-cant-analyze-an-sec-filing-researchers-find.html}}~\cite{islam2023financebench}.
To the best of our knowledge, this is the first systematic study on chunking for document understanding and more specifically for processing financial reports.

%

\section{Related work}

RAG is an innovative method that has emerged to enhance the performance of LLMs by incorporating external knowledge, thereby boosting their capabilities. This technique has undergone substantial research, examining various configurations and applications. Key research includes Gao et al.'s~\cite{gao2023retrieval} detailed analysis of RAG configurations and their role in enhancing Natural Language Processing (NLP) tasks, reducing errors, and improving factual accuracy. Several context retrieval methods are proposed to dynamically retrieve documents to improve the coherence of generated outputs~\cite{anantha2023context}. Other research introduced advancements in RAG, including reasoning chain storage and optimization strategies for retrieval, respectively, broadening the scope and efficiency of RAG applications in LLMs~\cite{lewis2020retrieval}. More recent work has compared RAG vs LLM fine-tuning, and identified that applying both improves the performance of each individual method~\cite{balaguer2024rag}.

Chunking has been identified as the key factor in the success of RAG, improving the relevance of retrieved content by ensuring accurate embedding of text with minimal noise. Various strategies have been developed for text subdivision, each with its unique approach. They can be summarized as follows: the \textbf{\textit{fixed size strategy}} divides text into uniform segments, but it often overlooks the underlying textual structure. In contrast, the \textbf{\textit{recursive strategy}} iteratively subdivides text using separators like punctuation marks, allowing it to adapt more fluidly to the content. The \textbf{\textit{contextual strategy}} takes this a step further by employing NLP techniques such as sentence segmentation to represent the meaning in context. Lastly, the \textbf{\textit{hybrid strategy}} combines different approaches, offering greater flexibility in handling diverse text types~\cite{pinecone_chunking}. However, an area yet to be explored in RAG chunking based on element types (document structure), which involves analyzing the inherent structure of documents, such as headings, paragraphs, tables, to guide the chunking process. Although chunking by Markdown and LaTeX comes closer to addressing element types, it's not the same in nature as a dedicated approach that directly considers document structure and element types for chunking, which could potentially yield more contextually relevant chunks.


Exploring the structure of financial reports is an exceptional area for establishing optimal principles for chunking. The intricate nature of document structures and contents has resulted in most of the work processing financial reports focusing on the identification of structural elements. Among previous work, we find El-Haj et al.~\cite{el2014detecting} and the FinTOC challenges~\cite{juge2019fintoc,bentabet2020financial,el2021financial} that have worked at the document structure level for UK and French financial reports.
Additionally, there is recent work that considers U.S. SEC reports, which includes DocLayNet~\cite{pfitzmann2022doclaynet} and more specifically with the report tables in FinTabNet~\cite{zheng2021global}.

On the side of financial models, there is work in sentiment analysis in finance~\cite{rizinski2023sentiment}, which includes the pre-training of specialised models such as FinBERT by Liu et al.~\cite{liu2021finbert}, which is a BERT based model pre-trained on large corpora including large collections of financial news collected from different sites and FinBERT by DeSola et al,~\cite{desola2019finbert} trained on Wikipedia, BookCorpus and U.S. SEC data.
Additional models include BloombergGPT~\cite{wu2023bloomberggpt}, FinGPT~\cite{yang2023fingpt} and Instruct-FinGPT\cite{zhang2023instructfingpt}.

More advance datasets in the financial domain include FinQA~\cite{chen2021finqa}, LLMWare~\cite{llmware2023rag}, ConFIRM~\cite{choi2023conversational} and TAT-QA~\cite{zhu2021tat} among others~\cite{chen2022convfinqa,shah2022flue,kaur2023refind} that have been prepared for retrieval and or Questions and Answering (Q\&A) tasks over snippets of financial data that includes tabular data, which has allowed methods on large language models to be tested on them~\cite{singh2023zero}.

Most of the previous work has focused on understanding the layout of financial documents or understanding specific snippets of existing reports with different levels of complexity, but there has not been much research in understanding financial report documents, except some more recent work that includes FinanceBench~\cite{islam2023financebench}, in which a set of questions about the content of financial reports are proposed that includes the evidence snippet.

More specifically on document chunking methods for RAG, there are standard approaches being considered such as chunking text into spans of a given token length (e.g. 128 and 256) or chunking based on sentences. Open source projects already allow simple processing of documents (e.g. Unstructured\footnote{\url{https://unstructured.io}}, Llamaindex\footnote{\url{https://www.llamaindex.ai}} or Langchain~\footnote{\url{https://www.langchain.com}}), without explicitly considering the table structure on which these chunking strategies are applied.

Even though different approaches are available, an exhaustive evaluation of chunking applied to RAG and specifically to financial reporting, except for some limited chunking analysis~\cite{islam2023financebench,retteter2023finance}, is non-existent. 
In our work, we compare a broad range of chunking approaches in addition to more simple ones and provide an analysis of the outcomes of different methods when asking questions about different aspects of the reports.

\section{Methods}

In this section, we present the chunking strategies that we have evaluated.
Before describing the chunking strategies, we present the RAG environment in which these strategies have been evaluated and the dataset used for evaluation.

\subsection{RAG setting for the experiments}

\begin{figure}[htb] 
  \centering
  \includegraphics[width=1.0\textwidth]{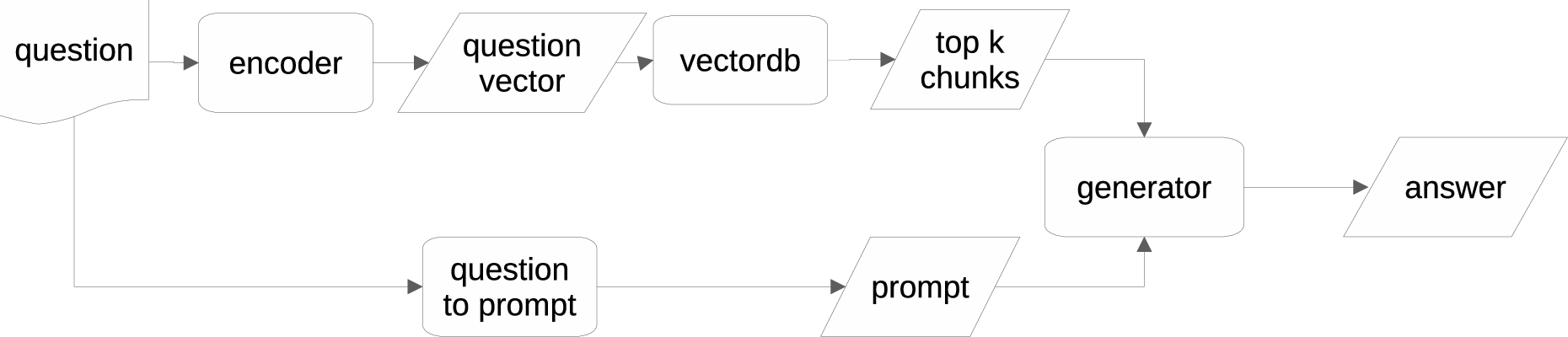}
  \caption{RAG steps to answer a question about a document}
  \label{fig:rag}
\end{figure}

The RAG pipeline used to process a user question is presented in figure~\ref{fig:rag} and is a common instance of a RAG~\cite{gao2023retrieval}.
Prior to answering any question about a given document, the document is split into chunks and the chunks are indexed into a vector database (vectordb).
When a question is sent to the RAG system, the top-k chunks most similar to the question are retrieved from the vector database and used to generate the answer using a large language model as generator.
In order to retrieve chunks from the vector database, the question is encoded into a vector that is compared to the vector previously generated from the chunks.
To prompt the generator, the question is converted into a set of instructions that instruct the LLM to find the answer within the top-k retrieved chunks.

In our experiments, we modify the way documents are~\emph{chunked} prior to being indexed in the vector database.
All other settings remain constant.
In the following sections, we describe in more detail each one of the components and processes used.

\subsection{Indexing and retrieval}

We have used the open source system Weaviate\footnote{\url{https://weaviate.io/developers/weaviate}} as our vector database.
As encoder model, we have used a sentence transformer~\cite{reimers-2019-sentence-bert} trained on over 256M questions and answers, which is available from the HuggingFace system\footnote{\url{https://huggingface.co/sentence-transformers/multi-qa-mpnet-base-dot-v1}}.

\begin{figure}[htb]  
  \centering  
  \includegraphics[width=1.0\textwidth]{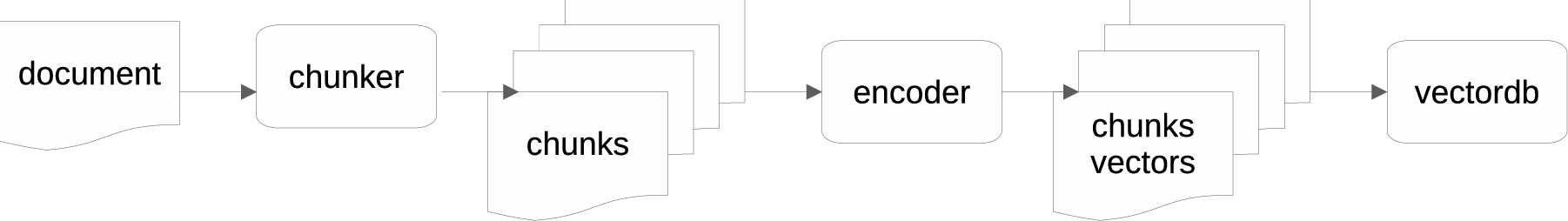}  
  \caption{Indexing of document chunks into the vector database}
  \label{fig:indexing}
\end{figure}

As shown in figure~\ref{fig:indexing}, to index a document, first the document is split into chunks, then each chunk is processed by an encoder model and then indexed into the vector database.
Based on the chunking strategy a document will be split into a larger or smaller set of chunks.

As shown in figure~\ref{fig:rag}, to retrieve chunks relevant to a question, the question is converted into a vector representation and the vector database returns a ranked list of chunks based on the similarity between question vector and the chunks in the database.
Weaviate implements an approximate nearest neighbours algorithm~\cite{malkov2018efficient} as their retrieval approach, which supports fast retrieval with high accuracy.
In our experiments, we retrieve the top-10 chunks for each question.

\subsection{Generation}

Once the vector database has retrieved the top-10 chunks based on a question, the generation module generates the answer.
To do so, a prompt based on the question and the retrieved chunks are provided to a large language model that generates the answer of the system.

We have used GPT-4~\cite{openai2023gpt4} as the generator, which has shown best performance compared to earlier versions. As well, its performance was better compared to existing open source alternatives~\cite{longchat2023} such as Mixtral~\cite{jiang2024mixtral}.
We used the prompt presented in figure~\ref{fig:example-prompt} that we designed on another similar RAG implementation with different document types.
The prompt conditions the answer to the query and the chunks, referred to as~\emph{source}, and if the generator cannot answer it should return~\emph{No answer}.

\begin{figure}[htb]
\centering
\begin{mdframed}
\begin{verbatim}
please answer the question below by referencing the list of sources
provided after the question; if the question can not be answered just
respond 'No answer'. The sources are listed after "Sources:".

Question: {query}

Sources: {key} - {source}
...
\end{verbatim}
\end{mdframed}
\caption{Example prompt template used by the generator}
\label{fig:example-prompt}
\end{figure}

\subsection{Chunking}

As a baseline chunking method, we have split the documents into chunks of size $n$ tokens ($n \in \{128, 256, 512\}$).
As well, an aggregation of the output by the indexing of different chunking configurations has been considered.


In addition to chunking based on the number of tokens, we have processed the documents using computer vision and natural language processing to extract elements identified in the reports. The list of elements considered are provided by the Unstructured\footnote{\url{https://unstructured-io.github.io/unstructured/introduction.html\#elements}} open source library. From the set of processing strategies, we use Chipper, a vision encoder decoder\footnote{\url{https://huggingface.co/docs/transformers/model\_doc/vision-encoder-decoder}} model inspired by Donut~\cite{kim2021donut} to showcase the performance difference. The Chipper model outputs results as a JSON representation of the document, listing elements per page characterized by their element type.
Additionally, Chipper provides a bounding box enclosing each element on the page and the corresponding element text.

These elements are sometimes short to be considered as chunks, so to generate chunks from elements the following steps have been followed. Given the structure of finance reporting documents, our structural chunking efforts are concentrated on processing titles, texts, and tables. The steps to generate element-based chunks are:

\begin{itemize}
    \item if the element text length is smaller than 2,048 characters, a merge with the following element is attempted
    \item iteratively, element texts are merged following the step above till either the desired length is achieved, without breaking the element
    \item if a title element is found, a new chunk is started
    \item if a table element is found, a new chunk is started, preserving the entire table
\end{itemize}

After element-based chunks have been derived, three types of metadata are generated to enrich the content and support efficient indexing. The first two types, generated via predefined prompt templates with GPT-4, include: 1) up to 6 representative keywords of the composite chunk 2) a summarised paragraph of the composite chunk. The third type is
3) Naive representation using the first two sentences from a composite chunk (a kind of prefix) and in the case of tables, the description of the table, which is typically identified in the table caption.

\subsection{Dataset}

To evaluate the performance of the different chunking approaches, we have used the FinanceBench dataset~\cite{islam2023financebench}. FinanceBench is a new benchmarking dataset designed to assess the capabilities of LLMs in answering open-book financial questions. The questions collected are realistic and applicable to real-world financial scenarios and include complex questions that require computational reasoning to arrive at conclusive answers. 

This dataset is made of 150 instances with questions and answers from 84 unique reports.
The dataset does not include the source documents, which we have downloaded.
We were able to recover only 80 documents, which reduces the number of questions to 141 from the original 150.
The distribution of Unstructured elements  predictions are shown in table~\ref{tab:elements-distribution}.

\begin{table}[ht!]
\centering
\caption{Unstructured element types distribution for \emph{Chipper} predictions against documents in FinanceBench.}
\begin{tabular}{|l|c|c|}
\hline
\textbf{Element Type} &   \textbf{\emph{Chipper} Entities}\\
\hline
NarrativeText    & 61,780 \\
Title            & 29,664 \\
ListItem          & 33,054 \\
UncategorizedText  & 9,400 \\
Footer            & 1,026 \\
Table             & 7,700 \\
Header            & 3,959 \\
Image             & 26 \\
FigureCaption     & 54 \\
Formula           & 29 \\
Address         & 229 \\
\hline
\textbf{Total}  & \textbf{146,921} \\
\hline
\end{tabular}
\label{tab:elements-distribution}
\end{table}

Documents have a varying number of pages, spanning from 4 pages (FOOTLOCKER\_2022\_8K\_dated-2022-05-20) to 549 pages (e.g. PEPSICO\_2021\_10K), with an average of 147.34 with std 97.78 with a total of 11,787 pages combined.
Each instance contains a link to the report, the question, a question type 
, the answer and supporting evidence, with page number where the evidence is located in the document, that allows for a closer evaluation of the results.
Based on the page number, evidence contexts are located in different areas in the documents, ranging from the first page in some cases up to page 304 in one instance. The mean page number to find the evidence is 54.58 with a standard deviation of 43.66, which shows that evidence contexts to answer the questions are spread within a document. 
These characteristics make FinanceBench a perfect dataset for evaluating RAG. An example instance is available in table~\ref{tab:financebench-example}.


\begin{table}[ht!]
\caption{Example question from the FinanceBench dataset}
\begin{center}
\begin{tabular}{|l|p{10cm}|}
\hline
Field & Value \\
\hline
financebench\_id & financebench\_id\_00859 \\
\hline
doc\_name & VERIZON\_2021\_10K \\
\hline
doc\_link & https://www.verizon.com/about/sites/default/files/2021-Annual-Report-on-Form-10-K.pdf \\
\hline
question\_type & 'novel-generated' \\
\hline
question & Among all of the derivative instruments that Verizon used to manage the exposure to fluctuations of foreign currencies exchange rates or interest rates, which one had the highest notional value in FY 2021? \\
\hline
answer & Cross currency swaps. Its notional value was \$32,502 million., \\
\hline
evidence\_text & Derivative Instruments We enter into derivative transactions primarily to manage our exposure to fluctuations in foreign currency exchange rates and interest rates. We employ risk management strategies, which may include the use of a variety of derivatives including interest rate swaps, cross currency swaps, forward starting interest rate swaps, treasury rate locks, interest rate caps, swaptions and foreign exchange forwards. We do not hold derivatives for trading purposes. The following table sets forth the notional amounts of our outstanding derivative instruments: (dollars in millions) At December 31, 2021 2020 Interest rate swaps \$ 19,779 \$ 17,768 Cross currency swaps 32,502 26,288 Forward starting interest rate swaps 1,000 2,000 Foreign exchange forwards 932 1,405 \\
\hline
page\_number & 85 \\
\hline
\end{tabular}
\end{center}
\label{tab:financebench-example}
\end{table}

\section{Results}

In this section, we evaluate the different chunking strategies using the FinanceBench dataset. Our evaluation is grounded in factual accuracy, which allows us to measure the effectiveness of each configuration by its precision in retrieving answers that match the ground truth, as well as its generation abilities.

We are considering 80 documents and 141 questions from FinanceBench.
Using the OpenAI tokenizer from the model~\emph{text-embedding-ada-002} that uses the tokenizer~\emph{cl100k\_base}\footnote{\url{https://platform.openai.com/docs/guides/embeddings/limitations-risks}}, there are on average 102,444.35 tokens with std of 61,979.45, which shows the large variability of document lengths as seen by the different number of pages per document presented above.

\subsubsection{Chunking Efficiency}

The first thing we analyzed is the total number of chunks, as it impacts indexing time. We would like to observe the relationship between accuracy and total chunk size. Table~\ref{tab:chunking-processing-statistics} shows the number of chunks derived from each one of the processing methods. Unstructured element-based chunks are closer in size to Base 512, and as the chunk size decreases for the basic chunking strategies, the total number of chunks increases linearly.

\begin{table}[ht!]
\caption{Chunks statistics for basic chunking elements and Unstructured elements}
\begin{center}
\begin{tabular}{|l|c|c|c|}
\hline
Processing & total chunks & mean chunks per document (std) & tables mean (std) \\
\hline
Base 128   & 64,058 & 800.73 (484.11) & N/A \\
Base 256   & 32,051 & 400.64 (242.04) & N/A \\
Base 512   & 16,046 & 200.58 (121.01) & N/A \\
\hline
Chipper  & 20,843 & 260.57 (145.80) & 96.20 (57.53)\\
\hline
\end{tabular}
\end{center}
\label{tab:chunking-processing-statistics}
\end{table}

\subsubsection{Retrieval Accuracy}

Secondly, we evaluate the capabilities of each chunking strategy in terms of retrieval accuracy. We use the page numbers in the ground truth to calculate the page-level retrieval accuracy, and we use ROUGE~\cite{lin2004rouge} and BLEU~\cite{papineni2002bleu} scores to evaluate the accuracy of paragraph-level retrieval compared to the ground truth evidence paragraphs.

As shown in Table~\ref{tab:rag-results-retrieval}, when compared to Unstructured element-based chunking strategies, basic chunking strategies seem to have higher page-level retrieval accuracy but lower paragraph-level accuracy on average. Additionally, basic chunking strategies also lack consistency between page-level and paragraph-level accuracy; higher page-level accuracy doesn't ensure higher paragraph-level accuracy. For example, Base 128 has the second highest page-level accuracy but the lowest paragraph-level scores among all. On the other hand, element-based chunking strategies showed more consistent results. 

A fascinating discovery is that when various chunking strategies are combined, it results in enhanced retrieval scores, achieving superior performance at both the page level (84.4\%) and paragraph level (with ROUGE at 0.568\% and BLEU at 0.452\%). This finding addresses an unresolved question: how to improve the accuracy of RAG.


The element based method provides the highest scores and it also provides a mechanism to chunk documents without the need to fine tune hyper-parameters like the number of tokens in a chunk. This suggests the element based method is more generalizable and can be applied to new types of documents. 

\subsubsection{Q\&A Accuracy}

Third, we evaluate the Q\&A accuracy for the chunking strategies. In addition to manual evaluation, we have investigated an automatic evaluation using GPT-4.
GPT-4 compares how the answers provided by our method are similar to or different from the FinanceBench gold standard, similar approaches have been previously evaluated~\cite{hada2023large,li2023evaluation,moore2023assessing,naismith2023automated}. The automatic evaluation allows scaling the evaluation efforts for the different chunking strategies that we have considered.
We used the prompt template in figure~\ref{fig:evaluation-template}.

\begin{figure}[htbp]
\centering
\begin{mdframed}
\begin{verbatim}
Begin with True or False. Are the two following answers (Answer 1 and
Answer 2) the same with respect to the question between single quotes
'{question}'?

Answer 1: '{ground_truth_answer}'
Answer 2: '{generated_answer}'
\end{verbatim}
\end{mdframed}
\caption{Evaluation prompt template. The $\{question\}$, $\{ground\_truth\_answer\}$ and $\{generated\_answer\}$ fields are substituted for each question accordingly.}
\label{fig:evaluation-template}
\end{figure}

Results in table~\ref{tab:rag-results} show that element-based chunking strategies offer the best question-answering accuracy, which is consistent with page retrieval and paragraph retrieval accuracy. 

Lastly, our approach stands out for its efficiency. Not only is element-based chunking generalizable without the need to select the chunk size, but when compared to the aggregation results that yield the highest retrieval scores. Element-based chunking achieves the highest retrieval scores with only half the number of chunks required compared to methods that do not consider the structure of the documents (62,529  v.s. 112,155). This can reduce the indexing cost and improve query latency because there are only half as many vectors to index for the vectordb that stores the chunks. This underscores the effectiveness of our solution in optimizing the balance between performance and computational resource requirements.



\begin{table}[ht!]
\caption{Retrieval results. For each chunking strategy, we show the number of chunks for all the documents (Total Chunks), Page Accuracy, and ROUGE and BLEU scores. ROUGE and BLEU are calculated as the maximum score from the list of recovered contexts for a question when compared to the known evidence for that question.}
\begin{center}
\begin{tabular}{|l|c|c|c|c|c|}
\hline
Chunking strategy & Total Chunks & Page Accuracy  & ROUGE   & BLEU \\
\hline
Base 128       & 64,058 & 72.34 &  0.383 &  0.181 \\
Base 256       & 32,051 & 73.05 & 0.433 &  0.231 \\
Base 512       & 16,046 & 68.09 & 0.455 & 0.250\\
\hline
Base Aggregation & 112,155 & 83.69 & 0.536 & 0.277 \\
\hline

\hline
Keywords Chipper  & 20,843 & 46.10 & 0.444 & 0.315\\
Summary Chipper & 20,843 & 62.41 & 0.473 & 0.350 \\
Prefix \& Table Description Chipper& 20,843 & 67.38 & 0.514 & 0.400\\
\hline
Chipper Aggregation & 62,529 & \textbf{84.40} & \textbf{0.568} & \textbf{0.452} \\
\hline
\end{tabular}
\end{center}
\label{tab:rag-results-retrieval}
\end{table}

\begin{table}[ht!]
\caption{Q\&A results. We show the percentage of questions with no answer and as well the accuracy either estimated automatically using GPT-4 or manually.}
\begin{center}
\begin{tabular}{|l|c|c|c|c|c|}
\hline
Chunking strategy & No answer & GPT-4 & Manual \\
\hline
Base 128        & 35.46 & 29.08 & 35.46\\
Base 256        & 25.53 & 32.62 & 36.88 \\
Base 512        & 24.82 & 41.84 & 48.23 \\
\hline
Keywords Chipper & 22.70 & \textbf{43.97} & \textbf{53.19}\\
Summary Chipper & 17.73 & \textbf{43.97} & 51.77 \\
Prefix \& Table Description Chipper & 20.57 & 41.13 & \textbf{53.19}\\
\hline
\end{tabular}
\end{center}
\label{tab:rag-results}
\end{table}

\section{Discussion}

Results demonstrate the efficacy of our approach in utilizing structural elements for chunking, which has enabled us to attain state-of-the-art performance on Q\&A tasks within the FinanceBench dataset (accuracy of 50\% vs 53.19\%) when an index is created from document chunks and used for generation. This method, which we refer to as~\textit{element base chunking}, has shown to yield consistent results between retrieval and Q\&A accuracy. 

We have observed that using basic 512 chunking strategies produces results most similar to the Unstructured element-based approach, which may be due to the fact that 512 tokens share a similar length with the token size within our element-based chunks and capture a long context, but fail keep a coherent context in some cases, leaving out relevant information required for Q\&A.
This is further observed when considering the ROUGE and BLEU scores in table~\ref{tab:rag-results-retrieval}, where the chunk contexts for the baseline have lower scores.

These findings support existing research stating that the best basic chunk size varies from data to data~\cite{barnett2024seven}. 
These results show, as well, that our method adapts to different documents without tuning. Our method relies on the structural information that is present in the document's layout to adjust the chunk size automatically.


We have evaluated aggregating the output of different chunking methods in the retrieval experiments as sown in table~\ref{tab:rag-results-retrieval}.
Even though the aggregation seems to be effective for retrieval, the Q\&A exceeded the GPT-4 token limit, which resulted in a non-effective Q\&A solution using the selected model.

As well, we evaluated variations of the prompt used to generate the answers (see figure~\ref{fig:example-prompt}).
Re-ordering the retrieval context and the question, but results were not statistically different.
We experimented as well with variations of the verbs using in the prompt, e.g. changing~\textit{referencing} with~\textit{using}, which seemed to lower the quality of the answers generated.
This shows that prompt engineering is a relevant factor in RAG.










We evaluated using GPT-4 for evaluation instead of relying on manual evaluation.
In most cases, GPT-4 evaluated correctly but failed when a more elaborate answer is provided.
As shown in figure~\ref{fig:evaluation-gpt-4-fail}, the answer is 39.7\% while the estimated answer is 39.73\% but with a detailed explanation of the calculation.

\begin{figure}[htbp]
\centering
\begin{mdframed}
\begin{verbatim}
Question: 'What is Coca Cola's FY2021 COGS % margin? Calculate what
was asked by utilizing the line items clearly shown in the income
statement.'? 

Answer 1: '39.7%'
Answer 2: 'From the income statement referenced on page 60 of
COCACOLA_2021_10K_embedded.json, we can see that Coca Cola's total
revenue in FY2021 was $38,655 million and their cost of goods sold
(COGS) was $15,357 million. To calculate the COGS % margin, we divide
the COGS by the total revenue and multiply by 100:
(15,357 / 38,655) * 100 = 39.73%
So, Coca Cola's FY2021 COGS % margin was approximately 39.73%.'
\end{verbatim}
\end{mdframed}
\caption{Evaluation prompt template}
\label{fig:evaluation-gpt-4-fail}
\end{figure}


\section{Conclusions and Future Work}

Results show that our element based chunking strategy improves the state-of-the-art Q\&A for the task, which is achieved by providing a better chunking strategy for the processed documents.
We provide comparison with baseline chunking strategies that allow us to draw conclusions about different chunking methods.

As future work, we would like to perform a similar evaluation in other domains, e.g. biomedical, to understand how our findings apply outside financial reporting.
As well, we would like studying which additional element types and/or relation between elements would support better chunking strategies for RAG.
Furthermore, we would like to study the impact of RAG configuration and element type based chunking.

\bibliographystyle{splncs04}
\bibliography{bibliography}

\end{document}